\documentclass[sn-mathphys,Numbered]{sn-jnl}% Math and Physical Sciences Reference Style
\usepackage{graphicx}%
\usepackage{multirow}%
\usepackage{amsmath,amssymb,amsfonts}%
\usepackage{amsthm}%
\usepackage{mathrsfs}%
\usepackage[title]{appendix}%
\usepackage{xcolor}%
\usepackage{textcomp}%
\usepackage{manyfoot}%
\usepackage{booktabs}%
\usepackage{algorithm}%
\usepackage{algorithmicx}%
\usepackage{algpseudocode}%
\usepackage{listings}%
\usepackage{subcaption}%
\usepackage{makecell}%
\theoremstyle{thmstyleone}%
\theoremstyle{thmstyletwo}%

\theoremstyle{thmstylethree}%

\begin{document}

\title[Article Title]{An Improved Artificial Fish Swarm Algorithm for Solving the Problem of Investigation Path Planning}


\author*[1]{\fnm{Qian} \sur{Huang}}\email{huangqian@hhu.edu.cn}

\author[1]{\fnm{Weiwen} \sur{Qian}}\email{qianweiwen@hhu.edu.cn}
\equalcont{These authors contributed equally to this work.}

\author[1]{\fnm{Chang} \sur{Li}}\email{lichang@hhu.edu.cn}
\equalcont{These authors contributed equally to this work.}

\author[2]{\fnm{Xuan} \sur{Ding}}\email{dingxuan@hhu.edu.cn}
\equalcont{These authors contributed equally to this work.}

\affil[1]{\orgdiv{School of Computer and Software, Hohai University},  \city{Nanjing}, \postcode{211100},  \country{China}}

\affil[2]{\orgdiv{China Mobile Suzhou R\&D Center},  \city{Suzhou}, \postcode{215000},  \country{China}}


\abstract{Informationization is a prevailing trend in today's world. The increasing demand for infor-mation in decision-making processes poses significant challenges for investigation activities, particularly in terms of effectively allocating limited resources to plan investigation programs. This paper addresses the investigation path planning problem by formulating it as a multi-traveling salesman problem (MTSP). Our objective is to minimize costs, and to achieve this, we propose a chaotic artificial fish swarm algorithm based on multiple population differential evolu-tion (DE-CAFSA). To overcome the limitations of the artificial fish swarm algorithm, such as low optimization accuracy and the inability to consider global and local information, we incorporate adaptive field of view and step size adjustments, replace random behavior with the 2-opt opera-tion, and introduce chaos theory and sub-optimal solutions to enhance optimization accuracy and search performance. Additionally, we integrate the differential evolution algorithm to create a hybrid algorithm that leverages the complementary advantages of both approaches. Experi-mental results demonstrate that DE-CAFSA outperforms other algorithms on various public da-tasets of different sizes, as well as showcasing excellent performance on the examples proposed in this study.}




\maketitle

\section{Introduction}\label{sec1}

With the rapid advancement of informatization, the decision-making process across various domains heavily relies on information collection and analysis. In order to make informed decisions in a dynamic and competitive environment, businesses must possess a precise understanding of industry-related information. Similarly, governments require substantial information gathering to comprehensively grasp public sentiment and assess prevailing circumstances prior to decision-making. Furthermore, the accuracy of scientific research projects is contingent upon reliable data collection, and individuals often need to conduct thorough investigations before reaching a decision. Consequently, the demand for effective investigation activities has surged, prompting researchers to explore methods of optimizing investigation plans within the constraints of limited resources, such as funds and time. This has emerged as a prominent area of research focus.

This paper addresses the problem of optimizing investigation path planning with the objective of minimizing costs. To achieve cost reduction, a systematic analysis and planning of the various components comprising the investigation cost are essential. Subsequently, an optimal decision can be made based on experimental verification and comparison. This behavior can be classified as an optimization problem, where the aim is to identify the most efficient approach.

Optimization problems typically employ mathematical methods to efficiently arrange, filter, and combine events within specific constraints. The objective is to find the maximum or minimum solution given the defined constraints. Commonly used optimization algorithms include Random Search \cite{Wang2013} (RS), Tabu Search \cite{Zhang2004} (TS), Simulated Annealing \cite{2017Improved,Kumar2017,Lan2016} (SA), Evolutionary Algorithm  (EA), Swarm Intelligence \cite{swarm2021} (SI), and others. Among these, the SA algorithm exhibits robustness in terms of initial values, universality, and ease of implementation, although it may have a longer optimization process. The TS algorithm, on the other hand, is a global iterative optimization algorithm that excels in local search capabilities but heavily relies on the initial solution.

Optimization theory finds broad applications in various domains. Problems such as the Traveling Salesman Problem \cite{Cook2012} (TSP), Clustering Problem \cite{2019Research}, Graph Partitioning, and Knapsack Problem \cite{2010Artificial} frequently involve the utilization of optimization theories. For instance, Rico et al. \cite{Rico2021} employed a modified version of a metaheuristic optimization algorithm to address the TSP problem in smart cities, yielding promising outcomes. Similarly, SánchezDíaz et al. \cite{app2021} proposed a straightforward yet highly effective hyper-heuristic model that significantly enhances the solution process for knapsack problems.

The Traveling Salesman Problem (TSP) is a well-known combinatorial optimization problem. It involves finding the shortest possible route that visits all given cities, where each city is visited only once, based on the given city distances. The investigation path planning problem addressed in this paper can be abstracted as a multi-objective optimization variant of the multi-traveling salesman problem. In this context, the objective is to optimize multiple objectives simultaneously while planning the investigation path, considering factors such as cost, time, or resource allocation.

The TSP problem has been approached through various solutions. Among probabilistic optimization algorithms, swarm intelligence algorithms have gained significant attention due to their versatility, parallelism, and flexibility. By studying the behavior, characteristics, and interactions of biological systems and simulating them using computer technology, swarm intelligence algorithms offer new principles and design ideas to solve optimization problems across different engineering and technical fields. This article focuses on researching and enhancing the artificial fish swarm algorithm within the realm of swarm intelligence heuristic algorithms. Swarm intelligence algorithms are currently a prominent area of research in the field of artificial intelligence. Existing research in this field includes Genetic Algorithm \cite{Alipour2018AHA} (GA), Ant Colony Optimization \cite{Gulcu2018} (ACO), Particle Swarm Optimization \cite{Kefi2016} (PSO), Artificial Fish Swarm Algorithm \cite{Huang2017} (AFSA), Grey Wolf Algorithm \cite{Oudira2019} (GWA), Firefly Algorithm \cite{yang2020firefly} (FA), etc.

Despite their advantages, previous swarm intelligence algorithms have exhibited certain limitations. Some algorithms suffer from slow convergence speeds, a heavy reliance on initial values, and a tendency to converge to local optima (e.g., ACO). Others face challenges in achieving a balance between local and global search performance, resulting in lower optimization accuracy (e.g., AFSA, PSO). Additionally, the reliance on a single optimization algorithm can lead to optimization bottlenecks. Consequently, many scholars have taken a keen interest in improving swarm intelligence algorithms. Sun et al. \cite{2014Research} introduced a modified version of AFSA that utilizes a novel piecewise adaptive function capable of producing different effects for different optimization objectives. Yin et al. \cite{yin2020} proposed the Cyber Firefly Algorithm (CFA), which combines the Firefly Algorithm (FA) with Glowworm Swarm Optimization (GSO) and Adaptive Memory Programming (AMP). CFA has demonstrated strong performance in global optimization of benchmark functions. These advancements showcase the ongoing efforts to enhance swarm intelligence algorithms and address their limitations.

To address the limitations of swarm intelligence algorithms in solving TSP problems, we have made significant improvements to the artificial fish swarm algorithm and introduced a novel approach known as the Chaotic Artificial Fish Swarm Algorithm based on Multiple Population Differential Evolution (DE-CAFSA). Our method incorporates adaptive field of view and step size adjustments, and introduces additional behaviors such as the 2-opt operation, chaos theory, and sub-optimal solutions to enhance the accuracy of the optimization process. Moreover, to overcome the limitations of relying on a single optimization algorithm, we have integrated the differential evolution algorithm, resulting in a new hybrid algorithm. To evaluate the effectiveness of our proposed algorithm, we conducted comparative experiments. These experiments provide a comprehensive assessment of the algorithm's performance and its ad-vantages over existing approaches.

The remaining sections of this document are structured as follows: Section 2 provides an introduction to the basic artificial fish swarm algorithm. Section 3 describes our proposed method, the Chaotic Artificial Fish Swarm Algorithm based on Multiple Population Differential Evolution (DE-CAFSA), in detail. Section 4 evaluates the performance of DE-CAFSA on a general dataset and includes a specific example to verify its effectiveness. Finally, Section 5 presents the conclusions drawn from our study.

\section{Related Methods}\label{sec2}

Swarm intelligence algorithms, as a significant branch of intelligent computing, have gained considerable attention from researchers due to their high-efficiency optimization speed and reduced reliance on problem-specific information. In this paper, we present an enhanced algorithm derived from the artificial fish swarm algorithm, which belongs to the realm of swarm intelligence algorithms. The artificial fish swarm algorithm is introduced below.

The Artificial Fish Swarm Algorithm (AFSA) \cite{Huang2017} is a bionic optimization algorithm, building upon research on animal swarm intelligent behavior. This algorithm emulates the foraging behavior of fish schools, leveraging the understanding that the areas with the highest fish concentration in water generally correspond to regions rich in nutrients. AFSA revolves around three fundamental behaviors observed in fish: preying, clustering, and following. By employing a top-down optimization model, the algorithm initiates from individual-level behaviors and proceeds with local optimization of each fish in the school. This collective effort aims to achieve a prominent global optimum for the entire group.

\begin{figure}[h]%
\centering
\includegraphics[width=0.5\textwidth]{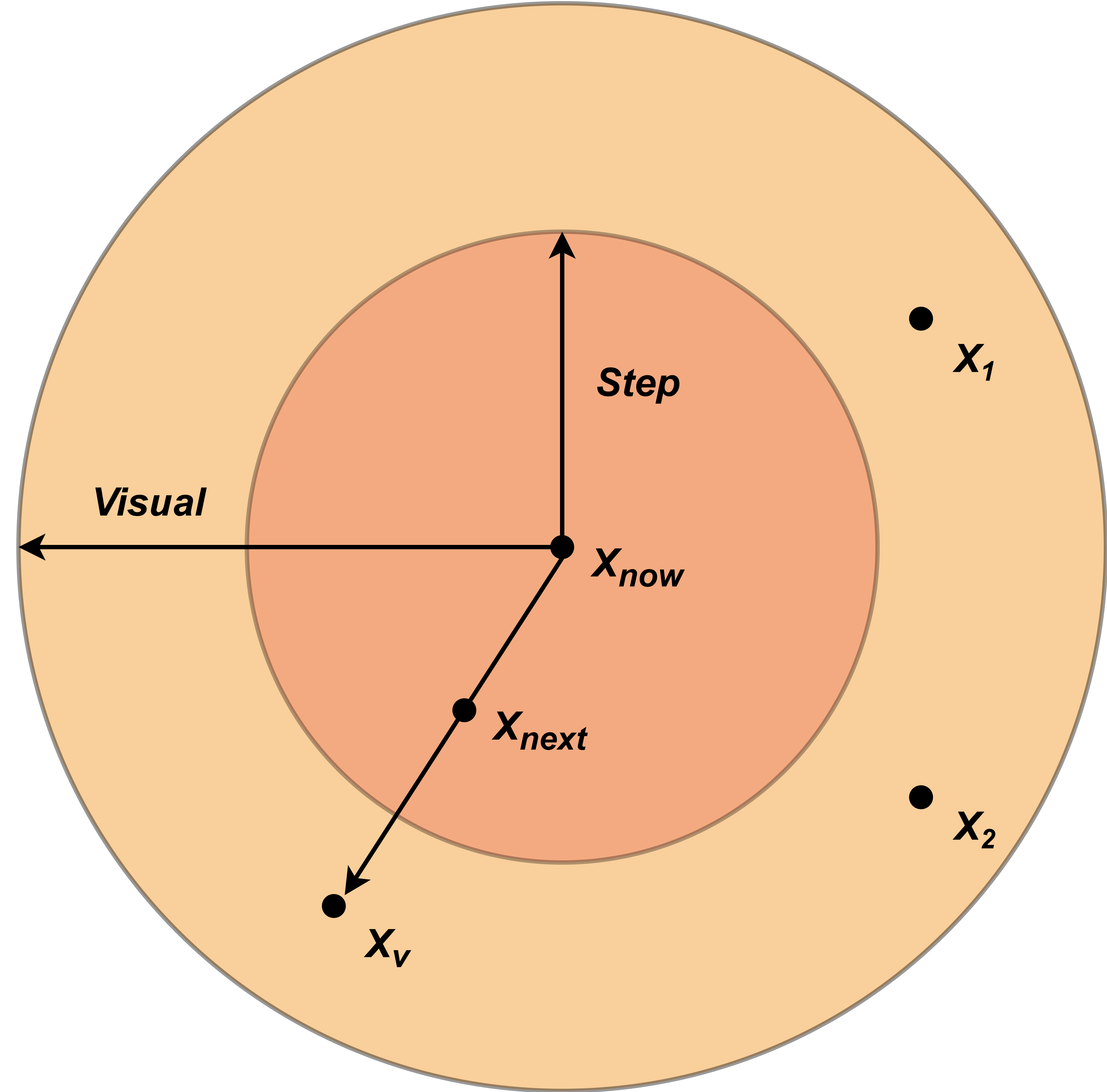}
\caption{Artificial fish abstract model}\label{fig1}
\end{figure}

The artificial fish's external perception is facilitated through its visual capabilities. Figure 1 illustrates the virtual vision of the artificial fish model. The field of view is denoted as Visual, while the current position of the artificial fish is denoted as $X_{now}$. The function $Rand()$ generates a random number between 0 and 1, and $Step$ represents the step size, within which the artificial fish can move in any direction. Typically, when the target direction offers a more favorable state compared to the current state, the artificial fish will move one step towards it. The status of each behavior is described below.

\subsection{Preying Behavior}\label{subsec21}
In the Preying Behavior, the artificial fish randomly selects another state within its perception range, as shown in Equation (1). If the newly selected state has a higher objective function value than the current state, the artificial fish moves one step closer to it, as shown in Equation (2). If not, the artificial fish continues to re-select states until a better option is found or until the maximum number of selections, $Trynum$, is reached. If no better state is found, the artificial fish moves one step randomly, as shown in Equation (3). This iterative process enables the artificial fish to actively search for and approach more favorable states, continually adapting its position based on the objective function evaluation.

\begin{equation}
X_j = X_i + Visuanl*Rand()
\label{eq1}
\end{equation}

\begin{equation}
X_i^{(t+1)} = X_i^{(t)} + \frac{X_j - X_i^{(t)}}{\vert\vert X_j - X_i^{(t)} \vert\vert }*Step*Rand()
\label{eq2}
\end{equation}

\begin{equation}
X_i^{(t+1)} = X_i^{(t)} + Step*Rand()
\label{eq3}
\end{equation}

Equation (1) describes the process in which the artificial fish, denoted as $X_i$, randomly selects a state $X_j$ within its field of view. Equation (2) then compares the objective function values $Y_i$ and $Y_j$ associated with $X_i$ and $X_j$, respectively. If $Y_j$ is found to be superior to $Y_i$, the artificial fish $X_i$ adjusts its position by moving one step in the direction of $X_j$. Where $X_i^{(t)}$ and $X_i^{(t+1)}$  represent the current and next positions of the i-th artificial fish.

\subsection{Clustering Behavior}\label{subsec22}

The artificial fish engages in an exploration process by assessing the number of partners within its current neighborhood. It then calculates the center position based on the positions of these partners. If the objective function value of the center position is superior to the current position and the center position is not densely populated, the artificial fish adjusts its position by moving one step towards the center position. However, if the objective function value of the center position is not better or if the center position is densely populated, the artificial fish switches to the foraging behavior. This behavior allows the artificial fish to strategically navigate towards more promising positions when conditions are favorable, while resorting to foraging when an advantageous center position is not available or when the environment is crowded.

\begin{equation}
X_i^{(t+1)} = X_i^{(t)} + \frac{X_c - X_i^{(t)}}{\vert\vert X_c - X_i^{(t)} \vert\vert }*Step*Rand()
\label{eq4}
\end{equation}

Equation (4) represents the process in which the artificial fish $X_i$ examines the number of partners within its current field of view and calculates the center position, denoted as $X_c$. If the center position of the partners is found to be in a superior state and not excessively crowded, the artificial fish $X_i$ adjusts its position by moving one step towards the center position. 

\subsection{Following Behavior}\label{subsec23}
In the Following Behavior, the artificial fish investigates the optimal position among its neighboring fish. If the objective function value of the optimal position surpasses that of the current position and the optimal neighbor is not excessively crowded, the artificial fish adjusts its position by moving one step towards the optimal neighbor. However, if the objective function value of the optimal position is not better or if the optimal neighbor is heavily populated, the artificial fish switches to foraging behavior.

\begin{equation}
X_i^{(t+1)} = X_i^{(t)} + \frac{X_j - X_i^{(t)}}{\vert\vert X_j - X_i^{(t)} \vert\vert }*Step*Rand()
\label{eq5}
\end{equation}

Equation (5) represents the process in which the artificial fish $X_i$ explores its current field of view to identify the optimal partner $X_j$ with the highest objective function value, denoted as $Y_j$. If the state of the optimal partner is found to be superior to the current state of $X_i$ and the environment is not overly crowded, $X_i$ adjusts its position by moving one step towards the optimal partner.

\subsection{Random Movement Behavior}\label{subsec24}
In the Random Movement Behavior, which is a default behavior of foraging, the artificial fish randomly moves within its field of view. Equation (6) describes the process in which the artificial fish $X_i$ randomly adjusts its position by taking one step, leading to a new state. This behavior allows the artificial fish to explore its environment without any specific guidance or objective, potentially discovering new states or paths through random movements.

\begin{equation}
X_i^{(t+1)} = X_i^{(t)} + Visuanl*Rand()
\label{eq6}
\end{equation}

The artificial fish swarm algorithm has been successfully applied in various fields due to its versatility. In the field of robot path planning, Zhang et al. \cite{2016The} proposed an improved artificial fish swarm algorithm for robot obstacle avoidance. Wang et al. \cite{WANG2017Improved} utilized AFSA with cross mutation for multi-threshold image segmen-tation in image processing, resulting in improved segmentation quality and reduced information loss. In neural networks, He et al. \cite{He2015} applied AFSA to train the weights and thresholds of BP neural networks, achieving faster convergence and improved accuracy. Additionally, in solving the TSP problem, AFSA has demonstrated better performance compared to standard particle swarm algorithms and basic genetic algorithms. These applications highlight the effectiveness and advantages of AFSA in various optimization and decision-making scenarios.

\section{Proposed Method}\label{sec3}
\subsection{Chaotic Artificial Fish Swarm Algorithm with Adaptive Field of View and Step Size(CAFSA)}\label{subsec31}
\subsubsection{Adaptive Field of View and Step Size}\label{subsec311}
The field of view plays a crucial role in determining the range of activities for artificial fish. In the original artificial fish swarm algorithm, the field of view parameter remains fixed. However, as the artificial fish approach the optimal solution, a majority of them tend to gather near the optimal solution. Nonetheless, using the original field of view for foraging purposes hinders the effectiveness of finding the best solution. When the initial field of view is set to a large value, the artificial fish exhibits strong global search capabilities, enabling rapid convergence in the early stages of the algorithm. However, the convergence speed gradually slows down in later stages due to the large field of view. The extensive field of view increases the number of trial at-tempts, leading to algorithmic complexity and slower search performance. On the other hand, if the initial field of view is set to a small value, the local search ability is enhanced, but this also increases the computational load and slows down the overall convergence speed of the algorithm, making it susceptible to getting stuck in local extrema.

Similarly, the step length determines the distance that an artificial fish moves at each iteration. In the basic artificial fish swarm algorithm, the step length is set as a constant value, which exhibits similar limitations as the field of view. If the step size is too large, it may lead to obstacles during foraging and result in unnecessary oscillations. Conversely, if the step size is too small, it affects the movement speed and may hinder the convergence of the algorithm.

According to this characteristic, we can adopt a strategy where the algorithm initially assigns a relatively large field of view and step size in the initial state. This al-lows the artificial fish to enhance its global optimization ability in a broad search space. As the algorithm progresses and the possible area of the optimal solution is roughly identified, the field of view and step size are gradually narrowed down. This adjustment strengthens the artificial fish's ability to optimize locally within the vicinity of the optimal solution, enabling a more focused search in that specific region.

In this paper, the field of view and moving step length are adjusted based on the following equation, which facilitates the adaptive refinement of the search process.

\begin{equation}
    \begin{cases}
    Visual(k+1) = (1 - \frac{k - 1}{K - 1})*Visual(k)   &;  1 - \frac{k - 1}{K - 1} \geq \beta \\
    Visual(k+1) = \beta * Visual(0) &;  1 - \frac{k - 1}{K - 1} \textless \beta, 
    \end{cases}
\label{eq7}
\end{equation}

\begin{equation}
    \begin{cases}
    Step(k+1) = (1 - \frac{k - 1}{K - 1})*Step(k)   &;  1 - \frac{k - 1}{K - 1} \geq \beta \\
    Step(k+1) = \beta * Step(0) &;  1 - \frac{k - 1}{K - 1} \textless \beta, 
    \end{cases}
\label{eq8}
\end{equation}

where $k$ is the current number of iterations, and the maximum number of iterations is $K$, $\beta \in (0,1)$ is the lower limit factor. $Visual(0)$ is the initial value of the field of view and $Step(0)$ is the initial value of the moving step.

This formula dynamically adjusts the weighting of the field of view and step size based on the current iteration number of the algorithm. It takes into account the balance between exploration and exploitation. If the field of view and step size become too small in the later stages of the algorithm, there is a risk of missing the global optimal solution. To prevent this, when the values of the field of view and step size reach a certain limit, their reduction is halted, and their values remain unchanged.

In Equation (7) and (8), the parameter $\beta$ can be utilized to flexibly control the minimum values of the field of view and step length, ensuring that they do not become excessively small. This provides a mechanism for balancing the exploration and exploitation abilities of the artificial fish swarm algorithm throughout the optimization process.

\subsubsection{Local Optimization Strategy}\label{subsec312}
In order to enhance the optimization capability of the algorithm, this study introduces the concept of the 2-opt optimization algorithm into the artificial fish swarm algorithm. When an artificial fish is unable to perform successful random movement during foraging behavior after reaching the maximum number of trials, the 2-opt operation is employed as an alternative. This operation involves reorganizing the artificial fish with bottleneck issues in foraging, aiming to achieve local optimization.

The 2-opt algorithm, also known as 2-exchange, is based on the idea of optimizing two elements at a time. Given a feasible tour, two edges are selected arbitrarily, and a new path is generated by gradually eliminating the crossover segments. If the new path is superior to the original path, it is replaced; otherwise, the original path is retained. This process allows for local optimization by iteratively improving the tour through pairwise edge exchanges.

\begin{figure}[h]%
\centering
\includegraphics[width=1\textwidth]{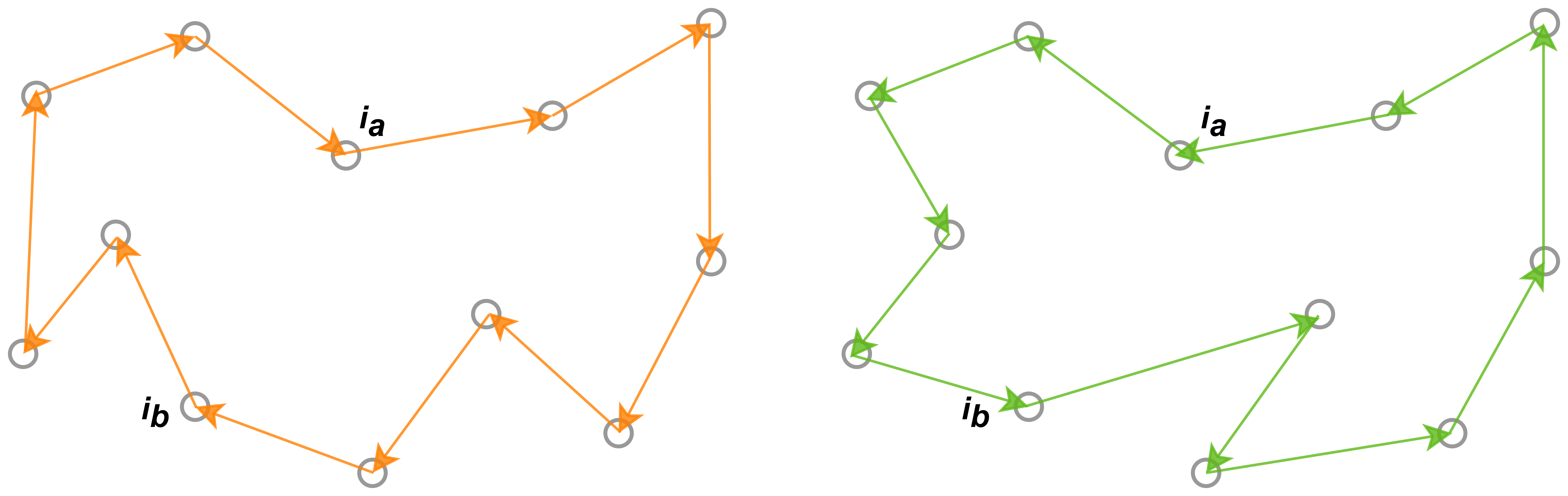}
\caption{2-opt path exchange}\label{fig2}
\end{figure}

Figure 2 illustrates an example of a 2-opt route exchange for a single traveling salesman problem. In this example, two points, $i_a$ and $i_b$, are selected from the tour loop of the traveling salesman's route. The positions of these two points are adjusted within the loop, and the path between these points is reversed, while the positions of the remaining points remain unchanged. This operation generates a new tour, which can be compared with the original tour to achieve local optimization. By iteratively applying such route exchanges, the algorithm aims to improve the quality of the tour by eliminating crossovers and finding shorter paths between selected points.

\subsubsection{Global Optimization Strategy}\label{subsec313}
To address the risk of getting trapped in local optima, a global optimization strategy is implemented in the algorithm. When an artificial fish engages in foraging behavior and fails to find a better state after reaching the maximum number of trials ($Trynum$), the 2-opt operation is performed. However, the 2-opt operation still carries the risk of converging to a local extreme value. In order to facilitate the algorithm's es-cape from local optima and achieve the global optimum more easily, Zhu et al. \cite{ZHU2015} introduced a novel foraging behavior that accepts sub-optimal solutions. This approach effectively lowers the foraging threshold throughout the optimization process, increasing the likelihood of escaping local extremes.

Despite the advancements mentioned above, the algorithm may encounter challenges associated with meaningless diffusion in the early stages. In the early stages, the algorithm has a larger state selection space, resulting in increased complexity and computation. This issue arises due to the phenomenon of individual state convergence, which tends to occur when the algorithm progresses into the later stages. Therefore, while efforts have been made to enhance the algorithm's ability to escape local optima, there is a need to strike a balance between exploration and exploitation to ensure efficient optimization without unnecessary diffusion.

To address the issue of meaningless diffusion in the early stages and strike a balance between exploration and exploitation, this paper introduces a restriction condition for expanding the foraging behavior. Specifically, the expansion of foraging behavior is only allowed when the current number of iterations reaches half of the maximum number of iterations ($k \geq 1/2K$). This restriction aims to prevent unnecessary diffusion and excessive complexity in the early stages of the algorithm when the state selection space is larger.

\subsubsection{Chaos Theory}\label{subsec314}
Chaos theory, characterized by seemingly chaotic and complex pseudo-random behavior, has been applied to practical engineering to address the issue of falling into local optima. By incorporating chaos theory into the algorithm, the sensitivity to initial conditions is enhanced. Even with very similar initial conditions, the motion trajectories exhibit significant divergence, capturing the essence of nonlinear phenomena.

One commonly used chaotic mapping is the logistic chaotic mapping, known for its simple form and robust sensitivity to initial conditions. The logistic mapping is represented by Equation (9):

\begin{equation}
   X_{(i+1)} = \mu X_i(1-X_i)
\label{eq9}
\end{equation}

Where $X_i \in (0,1)$ represents the state variable, and $i$ is the current iteration times of the algorithm. $\mu$ is the control parameter in the interval (0,4]. When $\mu = 4$, it is in a completely chaotic state.

This paper leverages logistic mapping from chaos theory to enhance the algorithm's global search capability. Based on this, the paper introduces a gradual expansion of the search range for artificial fish, enabling fine-grained exploration. In cases where searching within a small range fails to facilitate significant progress, the search space is expanded. The formula for generating a new state through chaotic search using the logistic map can be expressed as follows:

\begin{equation}
    \begin{cases}
     X_{new} = X_{best} + \Delta X\\
    \Delta X = -k * Step + 2k*Step*Z_i, 
    \end{cases}
\label{eq10}
\end{equation}
Where $X_{best}$ is the original state before search, Step is the moving step, k is the integer with initial value of 1, and $Z_i$ is the mapping variable.

\subsection{Chaotic Artificial Fish Swarm Algorithm Based on Multiple Population Differential Evolution(DE-CAFSA)}\label{subsec32}
\subsubsection{Multi-group Strategy}\label{subsec321}

To ensure independent evolution and avoid interference between sub populations, the multiple population strategy is employed. Each sub population follows its own mutation strategy, which introduces diversity among the sub populations. These differences enable the algorithm to explore multiple directions and expand the search range beyond the current narrow limits. Consequently, the algorithm's behavior becomes richer and more varied. In the same time interval, three sub populations evolve simultaneously, leading to a significant reduction in the time complexity of the algorithm. This parallel evolution allows for faster exploration and exploitation of the search space. This strategy is introduced in this section, specifically when the optimal state on the bulletin board remains unchanged after multiple iterations. At this point, the artificial fish swarm is randomly divided into three sub populations.

Assume that the overall population is $S$, The sub populations are $S_1,S_2,S_3$, and the total population size is $N$, which $N_i$  is the size of the sub population $S_i (i=1,2,3)$ and $\lambda_i$ is the proportion of sub population size randomly assigned. The division meet the following equation. Where $\lambda_i \in [0,1]$.

\begin{equation}
    \begin{cases}
     S = S_1 \bigcup S_2 \bigcup S_3\\
     N_i = \lambda_i*N\\
     \lambda_1 + \lambda_2 + \lambda_3 = 1
    \end{cases}
\label{eq11}
\end{equation}

\subsubsection{Differential Evolution Algorithms}\label{subsec322}
Differential Evolution (DE) algorithms \cite{Mingsheng2014Research,BENGUEDRIA2020366,CUI2020104181} are population-based evolutionary algorithms that rely on the competition and cooperation of individuals within the population. They have proven to be effective in solving complex global optimization problems. The key advantage of the DE algorithm lies in its simplicity and ease of integration with other algorithms, allowing for the construction of hybrid algorithms with improved performance and tailored requirements. The DE algorithm possesses unique memory capabilities, enabling it to assess the current search situation and dynamically adjust the search strategy. It demonstrates robustness and exhibits strong global convergence abilities. The fundamental concept of the DE algorithm involves randomly selecting two individual vectors from the initial population, multiplying the difference between the two vectors by a mutation operator, and adding the result to a third individual vector to create a new parameter vector. This new parameter vector is then combined with a pre-selected target vector, forming a test vector. Through a comparison of the evaluation function values, the individual vector with better results in the test vector is retained and carried forward to the next iteration, guiding the exploration from local optima towards the global optimum solution.

The mutation operation is a crucial component of the DE algorithm, as it introduces diversity and exploration into the population. Different mutation strategies can have varying effects on the algorithm's performance. In the context of the multi-group strategy, each subpopulation $(S_1,S_2,S_3)$ in the artificial fish swarm algorithm adopts a different mutation strategy from the differential evolution algorithm.

The subpopulation adopts the "DE/rand/1" strategy. This strategy utilizes a random individual as the mutation operator, which allows for an expanded search range and a good global search ability. Specifically, one artificial fish, denoted as $X_{r_1}$, is randomly selected from the current iteration's artificial fish swarm. Additionally, two other artificial fish, $X_{r_2 }$ and $X_{r_3 }$, are randomly chosen from the subpopulation $S_1$.To generate a variant individual, denoted as $V_{i/next}$, the difference between the states of $X_{r_2 }$ and $X_{r_3 }$ is scaled and added to the state of $X_{r_1}$ that was selected initially. This process is represented by Equation (12). The scaling factor F, which lies between the range of [0, 1], controls the speed of development for the population. Typically, a larger value of F enhances the global search ability, while a smaller value improves the local search ability. Take the " DE/RAND/1" strategy as an example to show the generation process of the variant individual $V_{i/next}$, as shown in Figure 3(a).

\begin{equation}
    V_{i/next} = X_{r_1} + F*(X_{r_2 } - X_{r_3 })
\label{eq12}
\end{equation}

The subpopulation $S_2$ in the artificial fish swarm algorithm adopts the "DE/best/1" strategy for mutation. In this strategy, the mutation operator is the best artificial fish in the current iteration, denoted as $X_{best}$. This approach provides strong guidance to the individuals in the population and guides them towards the global optimum. As shown in Equation (13).

\begin{equation}
    V_{i/next} = X_{best} + F*(X_{r_1 } - X_{r_2 })
\label{eq13}
\end{equation}

The sub population $S_3$  adopts the "DE/rand to best/1" strategy, which is similar to the "DE/rand/1" strategy but with a slight modification. The mutation operator is still randomly selected, but the difference vector is extended from one to two. Under the "DE/rand to best/1" strategy, the mutant individual is influenced not only by the state difference between the optimal fish $X_{best}$ and the current artificial fish $X_i$ after vector scaling but also by two randomly selected artificial fish, $r_1$ and $r_2$. This strategy accelerates the convergence speed of searching for the optimal solution in the later stages of the algorithm and effectively reduces the calculation time.

Overall, the "DE/rand to best/1" strategy combines the advantages of random selection and the influence of the best individual, allowing for a faster and more efficient search for the global optimum.

\begin{equation}
    V_{i/next} = X_i + K*(X_{best}-X_i) + F*(X_{r_1 } - X_{r_2 })  
\label{eq14}
\end{equation}

\begin{figure}[h]
  \centering
  \begin{subfigure}[b]{0.45\linewidth}
    \includegraphics[width=\linewidth]{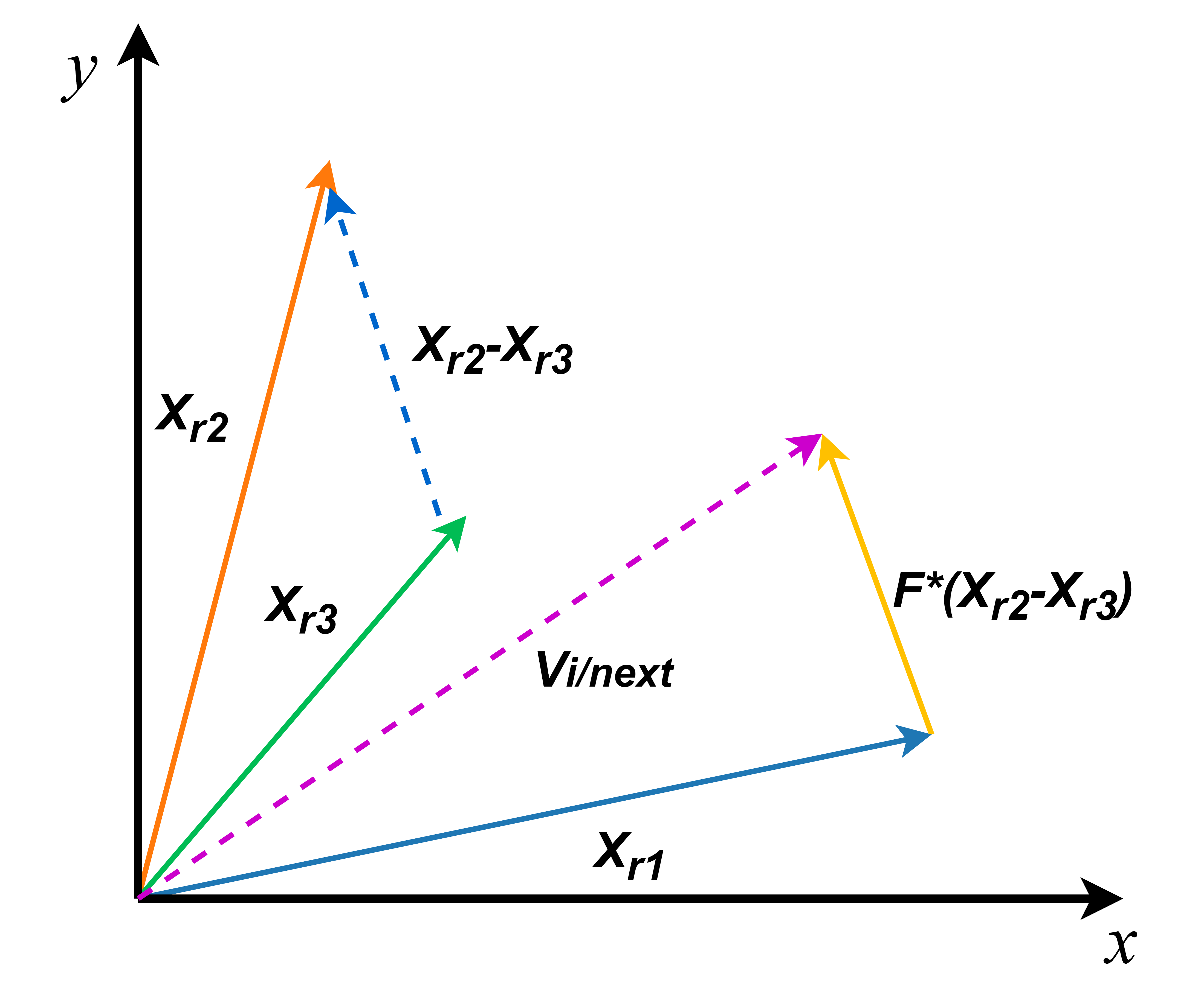}
    \caption{}
    \label{fig:sub1}
  \end{subfigure}
  \begin{subfigure}[b]{0.45\linewidth}
    \includegraphics[width=\linewidth]{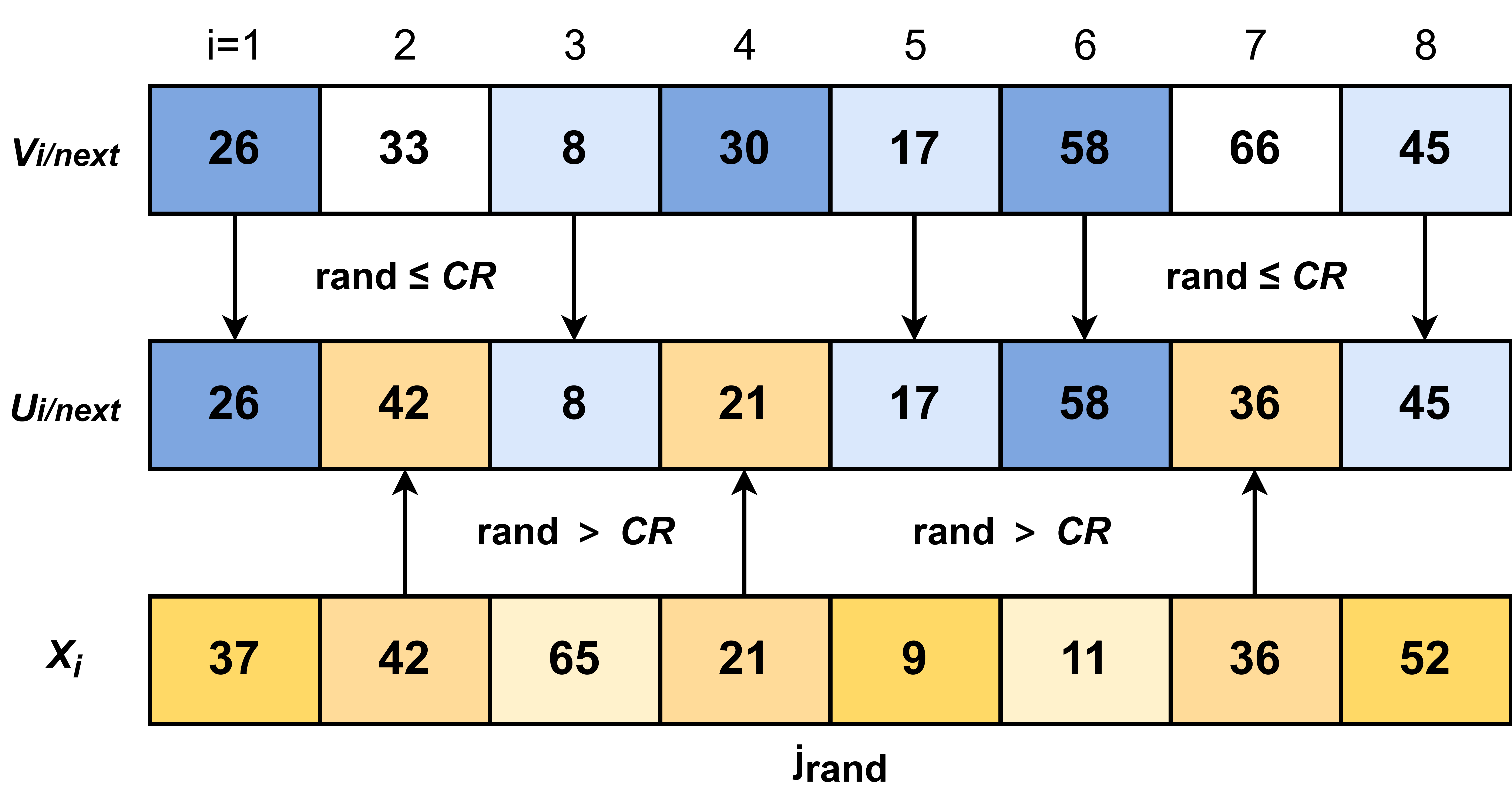}
    \caption{}
    \label{fig:sub2}
  \end{subfigure}
  \caption{Several operations of differential evolution algorithm(a) Schematic diagram of the generation process of mutant individuals; (b) Schematic diagram of binomial crossover operation.}
  \label{fig:fig}
\end{figure}

After the artificial fish swarm undergoes mutation, the mutated individuals are cross-mixed with the target individuals prior to mutation, resulting in the generation of a new artificial fish swarm denoted as $U_{i/next}$. The DE algorithm offers two crossover methods, namely binomial crossover and exponential crossover, with binomial crossover being the more widely used and superior approach. In the artificial fish swarm algorithm, the binomial crossover strategy is incorporated. The crossover probability parameter CR regulates the proportion of mutated individuals in the mixed population, typically controlled by itself within the range of [0,1]. The specific operation can be described using Equation (15):

\begin{equation}
    U_{i/next} = \begin{cases}
     V_{i/next,j}  &rand() \leq CR\ or\ j = j_{rand}\\
     X_{ij}        &       otherwise 
    \end{cases}
\label{eq15}
\end{equation}

To ensure the presence of individuals with variation in the new population and enhance the effectiveness of the binomial crossover operation while preventing algorithm stagnation, j takes on values from 1 to N, and $j_{rand}$ is a random positive integer selected from the range [1,N]. In the DE algorithm, $X_{ij}$ represents the state of the j-th individual in the current artificial fish swarm, specifically referring to a particular individual within the current population. On the other hand, $V_{i/next,j}$ represents the j-th individual among the newly generated individuals after the mutation operation, signifying a specific individual within the newly formed artificial fish swarm. The schematic diagram depicting the binomial crossover operation can be found in Figure 3(b).

The selection operation in the DE algorithm follows the commonly used "greedy" selection strategy, which enables offspring individuals to selectively inherit the states of superior individuals from their parents. The process involves calculating the fitness values of all the new individuals generated through the crossover operation. Each new individual's fitness value is then compared with the fitness values of the original individuals (i.e., individuals without mutation and crossover operations) one by one. Only when the fitness value of a new individual is better than that of the corresponding original individual, the new individual is retained. Otherwise, the original individual remains in the population. The selection operation can be represented by the following formula (16):

\begin{equation}
    X_{i/next} = \begin{cases}
     U_{i/next,j}  &f(U_{i/next,j}) \leq X_{ij}\\
     X_{ij}        &       otherwise 
    \end{cases}
\label{eq16}
\end{equation}

Among them, $U_{i/next,j}$ represents the j-th individual in the newly generated artificial fish school. The symbol $f()$ represents the fitness function, which is a mathematical function used to calculate the fitness value of an individual. The fitness function evaluates the individual's performance or suitability in the given problem space. It measures how well the individual solves the problem or achieves the desired objectives.

\subsubsection{Algorithm Steps}\label{subsec323}
The detailed steps of implementing the chaotic artificial fish swarm algorithm based on multiple population differential evolution (DE-CAFSA) are as follows:

Step 1: Artificial fish population and parameter initialization. Set the scale of the artificial fish swarm to $N$, the artificial fish field of view to $Visual$, the moving step length to Step, the crowding factor to $\delta$, the maximum number of trials to $Trynum$, the current number of iterations to $m$, and the maximum number of iterations to be $M$, $StopTime$ is the number of iterations during which the optimal record in the bulletin board does not change. Set the stagnation threshold to $MaxTime$, the scaling factor to $F$, and the crossover factor to $CR$.

Step 2: Initialize the bulletin board. Calculate the fitness value of all fish in the initial artificial fish swarm, and record the best individual on the bulletin board.

Step 3: Behavior selection. Artificial fish adopt the behavior strategy of variable field of view and variable step length to perform grouping behavior, rear-end collision behavior, and improved foraging behavior that accepts sub-optimal solutions with a certain probability. When performing the foraging behavior, if the optimal state cannot be obtained within the maximum number of trials, the artificial fish performs 2-opt local optimization. Select the current optimal state of the artificial fish and perform a chaotic search of Logistic mapping on the optimal state. If the new state is better, it will be retained, otherwise it will remain unchanged.

Step 4: Update the bulletin board. Compare the current optimal state with the record on the bulletin board. If the state is better than the bulletin board, replace it and set $StopTime=0$, otherwise $StopTime=StopTime+1$.

Step 5: Determine whether the number of stalls $StopTime$ reaches the stall threshold. If $StopTime \geq MaxTime$, it means that it has been reached. At this time, go to Step 6 to perform differential evolution operation on the artificial fish swarm; otherwise, go to Step 7.

Step 6: Differential evolution operation. The artificial fish swarm is randomly divided into three sub-populations using a multi-group strategy, and different mutation operations are applied to each sub-population, and then crossover and selection operations are performed, and then step four is performed.

Step 7: Determine whether the termination condition is reached. If the current iteration number $m$ is less than the maximum iteration number $M$, then $m=m+1$ and go to Step 3. Otherwise, it means that the algorithm is over and the optimal solution is output. The algorithm flow of DE-CAFSA is shown in Figure 4. 

\begin{figure}[h]%
\centering
\includegraphics[width=1\textwidth]{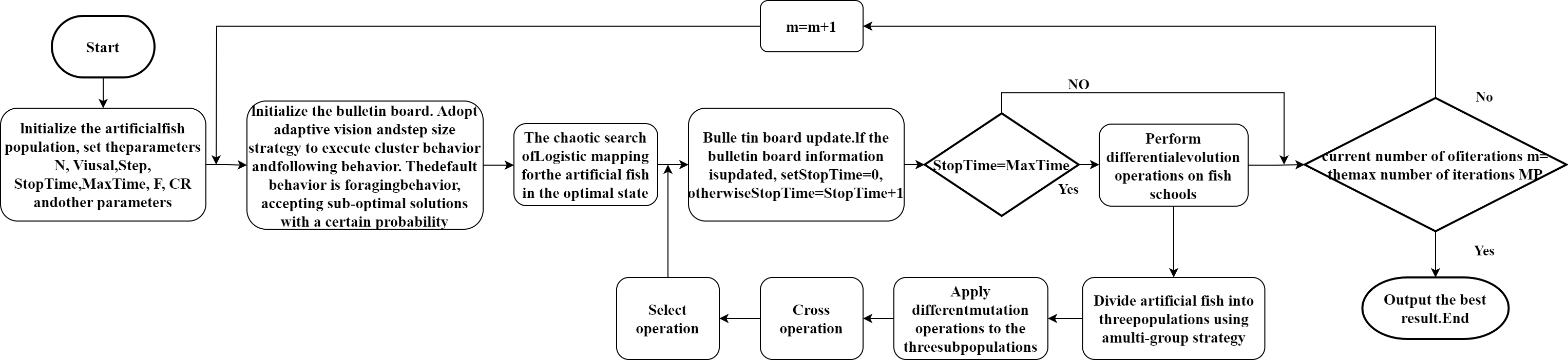}
\caption{DE-CAFSA flow chart}\label{fig5}
\end{figure}

\section{Experiments and Results}\label{sec4}
\subsection{Datasets and Experiment Settings}\label{subsec41}
We conducted simulation experiments to evaluate the effectiveness of the algorithm. The experiments were performed on a computer running the Windows 10 operating system, equipped with 8GB RAM and an i7-5500U CPU. The implementation code was written in MATLAB 2018a.
For the experiments, we utilized the "oliver30" dataset and the "eil101" dataset from the TSPLIB dataset, which are widely used benchmark datasets for the Traveling Salesman Problem. These datasets consist of 30 cities and 101 cities, respectively, allowing us to examine the algorithm's performance on small-scale and large-scale problem instances and assess its optimization capabilities.

To validate the effectiveness of the Adaptive Field of View and Step Size Chaos Artificial Fish Swarm Algorithm (CAFSA), we compared it with the Basic Artificial Fish Swarm Algorithm (AFSA), the Improved Genetic Algorithm (IGGA-SS), the Mon-key Group Algorithm based on 2-opt (2opt-IMA), and the Monkey Colony Algorithm (MK-IMA) that incorporates the Monkey King Decay mechanism. Through this comparative analysis, we aimed to demonstrate the superior performance of CAFSA.

Furthermore, to assess the effectiveness of the Mixed Operation (DE-CAFSA), we compared it with the Basic Differential Evolution Algorithm (DE), the Basic Artificial Fish Swarm Algorithm (AFSA), and the corresponding hybrid algorithm DE-AFSA. Building upon this comparison, we investigated the advantages of the novel hybrid algorithm DE-CAFSA, which combines the improved chaotic artificial fish swarm algorithm CAFSA and the DE algorithm mentioned earlier.

The following tables are the basic parameter settings of each algorithm. The parameter values of the algorithms not listed are similar to those of known algorithms:

\begin{table}[h]
\caption{Basic parameter setting of artificial fish swarm algorithm}\label{tab1}%
\begin{tabular}{@{}lll@{}}
\toprule
\textbf{Parameter Definition} & \textbf{Parameter Name}  & \textbf{Parameter Value} \\
\midrule
Population size    & $N$   & 20   \\
Maximum number of iterations    & $Max\_iter$   & 200    \\
Maximum number of trials    &  $Trynum$   & 20    \\
Visual field    &   $Visual$   & 10    \\
Moving step    &  $Step$   & 6    \\
Congestion factor   &   $\delta$   & 0.8    \\
\botrule
\end{tabular}
\end{table}

\begin{table}[h]
\caption{Basic parameter setting of genetic algorithm}\label{tab2}%
\begin{tabular}{@{}lll@{}}
\toprule
\textbf{Parameter Definition} & \textbf{Parameter Name}  & \textbf{Parameter Value} \\
\midrule
Population size    & $C$   & 20   \\
Number of iterations    & $Max\_iter$   & 2000    \\
Probability of choice    &  $P_s$   & 0.7    \\
Probability of crossover    &   $P_C$   & 0.7    \\
Probability of mutation    &  $P_m$   & 0.3    \\
\botrule
\end{tabular}
\end{table}

\begin{table}[h]
\caption{Basic parameter setting of differential evolution algorithm}\label{tab3}%
\begin{tabular}{@{}lll@{}}
\toprule
\textbf{Parameter Definition} & \textbf{Parameter Name}  & \textbf{Parameter Value} \\
\midrule
Population size    & $C$   & 20   \\
Number of iterations    & $Max\_iter$   & 2000    \\
Scaling factor    &  $F$   & 0.5    \\
Crossover probability    &   $CR$   & 0.5       \\
\botrule
\end{tabular}
\end{table}

\subsection{CAFSA Performance Analysis}\label{subsec42}

The experimental results of the CAFSA and its comparison algorithm on the two test sets are shown in Figure 5.

\begin{figure}[h]
  \centering
  \begin{subfigure}[b]{0.45\linewidth}
    \includegraphics[width=\linewidth]{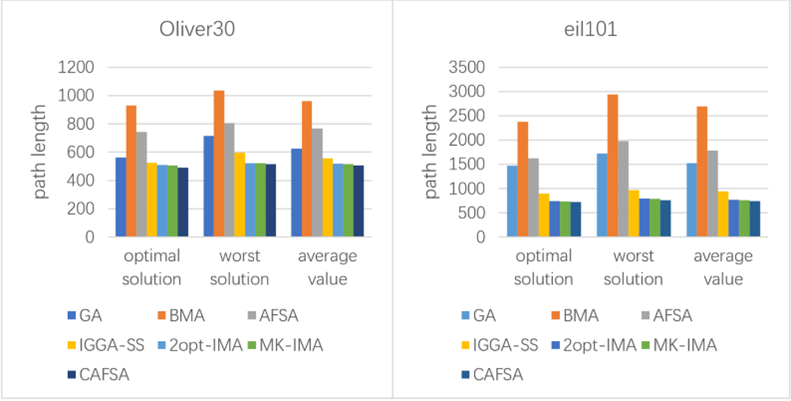}
    \caption{}
    \label{fig:sub1}
  \end{subfigure}
  \begin{subfigure}[b]{0.24\linewidth}
    \includegraphics[width=\linewidth]{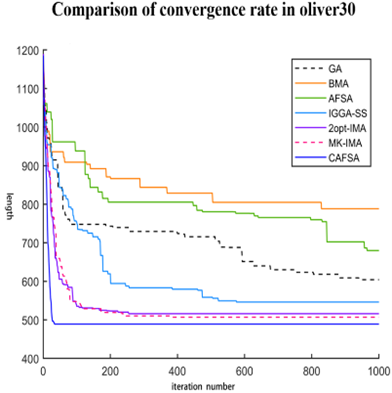}
    \caption{}
    \label{fig:sub2}
  \end{subfigure}
  \begin{subfigure}[b]{0.24\linewidth}
    \includegraphics[width=\linewidth]{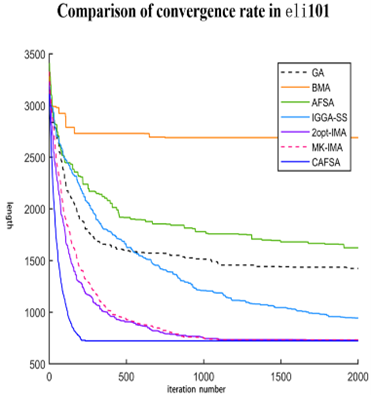}
    \caption{}
    \label{fig:sub3}
  \end{subfigure}
  \caption{Experimental results of CAFSA and other algorithms.}
  \label{fig:fig}
\end{figure}

The experimental results reveal that among the four improved algorithms, CAFSA exhibits a slight advantage in solution accuracy, indicating the effectiveness of parameter optimization, local optimization, and global optimization in the artificial fish swarm algorithm. Although CAFSA requires more time compared to the IGGA-SS algorithm, the trade-off between time and accuracy is favorable, with CAFSA achieving higher accuracy. When compared to the 2opt-IMA and MK-IMA algorithms, CAFSA demonstrates superior accuracy while incurring a lower time overhead. Consequently, it can be concluded that CAFSA not only offers high optimization precision and algorithm stability but also incurs minimal time overhead, thus showcasing notable ad-vantages in overall performance.

The convergence curves clearly demonstrate that the improved algorithms outper-form their corresponding basic algorithms in terms of convergence, across different scale test sets. Among the four improved algorithms, CAFSA exhibits the fastest con-vergence. It stands out as superior to the other three improved algorithms in terms of convergence performance. This indicates that the enhancements introduced in CAFSA have effectively accelerated the convergence rate, leading to more efficient optimiza-tion of the objective function.

\subsection{DE-CAFSA Performance Analysis}\label{subsec43}

The experimental results of the DE-CAFSA and its comparison algorithm on the two test sets are shown in Table 4 and Table 5.

\begin{table}[h]
\caption{The solution results of each algorithm on the test set Oliver30}\label{tab4}%
\begin{tabular}{@{}lllll@{}}
\toprule
\textbf{Algorithm} & \textbf{Optimal Solution}  & \textbf{Worst Solution}  & \textbf{Average Value} & \textbf{Average Time(s)}\\
\midrule
DE    & 638.5713   & 728.4548 & 678.8139 &  8.6\\
AFSA    & 743.3741   & 805.7888  & 766.7533 & 6.7 \\
DE-AFSA    &  564.7282   & 621.4936 & 590.9114 & 16.9   \\
CAFSA    &   489.0681   & 516.3028 & 505.9716 &  26.4  \\
DE-CAFSA    &  445.7660   & 513.3285 & 502.3212 & 30.6    \\

\botrule
\end{tabular}
\end{table}

\begin{table}[h]
\caption{The solution results of each algorithm on the test set eil101}\label{tab5}%
\begin{tabular}{@{}lllll@{}}
\toprule
\textbf{Algorithm} & \textbf{Optimal Solution}  & \textbf{Worst Solution}  & \textbf{Average Value} & \textbf{Average Time(s)}\\
\midrule
DE    & 2056.9085   & 2221.6874 & 2143.8102 &  29.3\\
AFSA    & 1623.5379	&1972.9826	&1784.5972	&42.8 \\
DE-AFSA    &  1422.5694	&1714.4281&	1681.4864	&47.6   \\
CAFSA    &  723.1209	&762.4521	&744.9705	&51.2  \\
DE-CAFSA    &  695.9502	&756.0367	&739.1097	&60.4   \\

\botrule
\end{tabular}
\end{table}

% \begin{table}[h]
% \caption{The solution results of each algorithm on the test set Oliver30}\label{tab4}%
% \begin{tabular}{@{}lllll@{}}
% \toprule
% \textbf{Algorithm} & \textbf{\thead{Optimal \\  Solution}}  & \textbf{\thead{Worst \\ Solution}}  & \textbf{\thead{Average \\ Value}} & \textbf{\thead{Average \\ Time(s)}}\\
% \midrule
% DE    & 638.5713   & 728.4548 & 678.8139 &  8.6\\
% AFSA    & 743.3741   & 805.7888  & 766.7533 & 6.7 \\
% DE-AFSA    &  564.7282   & 621.4936 & 590.9114 & 16.9   \\
% CAFSA    &   489.0681   & 516.3028 & 505.9716 &  26.4  \\
% DE-CAFSA    &  445.7660   & 513.3285 & 502.3212 & 30.6    \\

% \botrule
% \end{tabular}
% \end{table}

\begin{figure}[h]
  \centering
  \begin{subfigure}[b]{0.45\linewidth}
    \includegraphics[width=\linewidth]{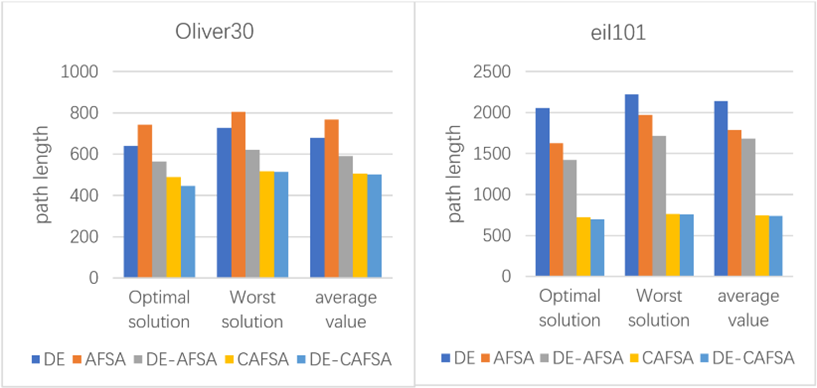}
    \caption{}
    \label{fig:sub1}
  \end{subfigure}
  \begin{subfigure}[b]{0.24\linewidth}
    \includegraphics[width=\linewidth]{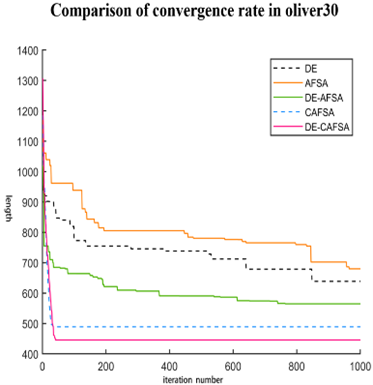}
    \caption{}
    \label{fig:sub2}
  \end{subfigure}
  \begin{subfigure}[b]{0.24\linewidth}
    \includegraphics[width=\linewidth]{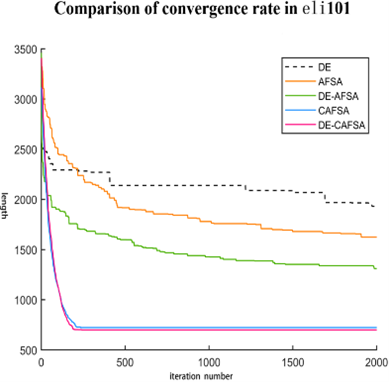}
    \caption{}
    \label{fig:sub3}
  \end{subfigure}
  \caption{Experimental results of DE-CAFSA and other algorithms.}
  \label{fig:fig}
\end{figure}

Comparing the average running time of various algorithms in Table 4 and Table 5, it can be seen that the hybrid algorithm runs longer than a single algorithm, and the more complex the algorithm consumes more time, but the overall operation The speeds are all within the acceptable range.

Figure 6 can intuitively show the solution effect of DE-CAFSA and other algorithms on the two test sets.

From the figure, it is evident that the hybrid algorithm consistently outperforms the individual algorithms, particularly when dealing with large-scale problems. The solution accuracy of DE-AFSA and DE-CAFSA is significantly better than that of the single algorithms. The merging of the improved chaotic artificial fish swarm algorithm CAFSA and the DE algorithm in DE-CAFSA results in higher accuracy than CAFSA alone when applied to different test sets. This demonstrates that the hybrid algorithm effectively overcomes the limitations of individual algorithms, leveraging their complementary advantages to achieve superior results.
Comparing the two hybrid optimization algorithms, DE-AFSA and DE-CAFSA, in terms of optimal solution, worst solution, and average value, it is clear that the new hybrid algorithm with the improved artificial fish swarm performs better than the basic hybrid algorithm. This comparison validates the enhanced accuracy and stronger ability to search for optimal solutions offered by the proposed hybrid algorithm.

The convergence curve analysis indicates that the scale variation of the test set has minimal impact on the convergence performance of each algorithm. The improved CAFSA and DE-CAFSA exhibit significantly enhanced convergence speed, displaying faster convergence in both test sets. Moreover, the hybrid algorithm DE-CAFSA does not lag behind the single algorithms in terms of convergence speed.

In Figure 7, the traveling path processes of the three groups (DE, AFSA, DE-CAFSA) when applied to the Oliver 30 and eil101 datasets are displayed. The traveling paths are reasonable, and the total path lengths are minimized, providing evidence of the superiority of the new hybrid algorithm DE-CAFSA proposed in this paper.

\begin{figure}[h]
  \centering
  \begin{subfigure}[b]{0.45\linewidth}
    \includegraphics[width=\linewidth]{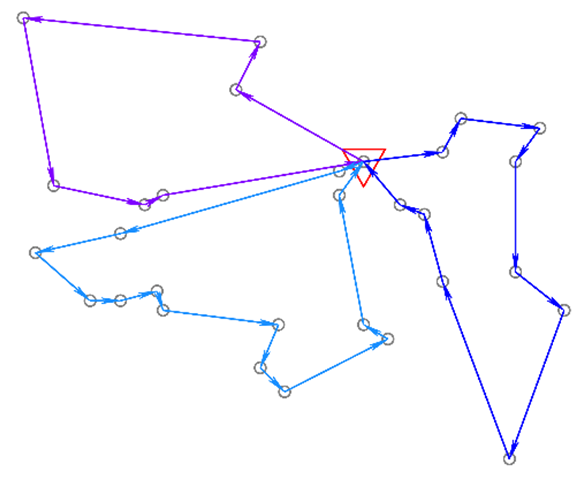}
    \caption{}
    \label{fig:sub1}
  \end{subfigure}
  \begin{subfigure}[b]{0.45\linewidth}
    \includegraphics[width=\linewidth]{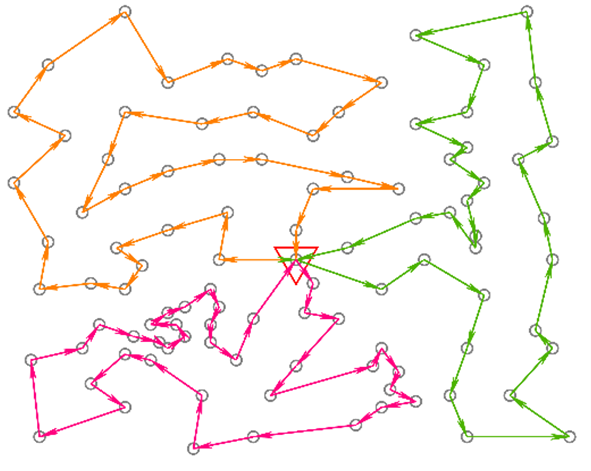}
    \caption{}
    \label{fig:sub2}
  \end{subfigure}
  \caption{Use DE-CAFSA to solve the path diagram of Oliver30 and eil101.}
  \label{fig:fig}
\end{figure}
 
\subsection{Modeling and Solving Practical Problems}\label{subsec44}
\subsubsection{Mathematical Modeling}\label{subsec441}
To verify the effectiveness of DE-CAFSA in solving the actual investigation path planning problem, a simulation will be conducted for planning the small watershed investigation path in Jiangsu Province. For this simulation, 100 small watersheds with relatively uniform distribution will be randomly selected as investigation sites. During the investigation process, the investigators will be divided into multiple groups. All groups will start from the same departure location and travel to other locations. Eventually, all groups need to return to the departure location. It is important to note that each location, except the departure location, can only be visited once by a specific group. Additionally, practical factors such as cost minimization need to be considered comprehensively when planning the driving route. The problem at hand can be formulated as a multi-objective optimization multi-traveling salesman problem within the context of the traveling salesman problem. A mathematical model will be established and solved to address this problem.

The investigation path planning problem aims to minimize the total cost, denoted as Cost. The total cost consists of three components: fuel cost ($C_1$), personnel cost ($C_2$), and other costs ($C_3$). The equation representing the total cost is as follows:

\begin{equation}
    Cost = C_1 + C_2 + C_3 
\label{eq17}
\end{equation}

(1) First establish the objective function of fuel cost. Fuel costs are controlled by two factors: path length and road type. Suppose there are $K$ research groups ($m$ people in each team) from the same starting point to $N$ planned locations to conduct investigation. If the k-th group is from site $i$ to site $j$, set $x_{ijk}=1$, otherwise it is equal to 0. The distance from site $i$ to site $j$ is represented by $d_{ij}$. Assuming that the total distance traveled by all groups is $D$, and the equation is as follows:

\begin{equation}
    D = \sum_{i=0}^{N} \sum_{j=0}^{N} \sum_{k=0}^{N}  x_{ijk}  d_{ij}
\label{eq18}
\end{equation}

In this example, the gasoline price is $P_1$ yuan per liter. $q$ represents the overall basic fuel consumption of the vehicle. $K_r$ is the road correction coefficient. In this example, referring to the GB/T 21010-2017 Land Use Status Classification Specification, the land status of the investigation site is divided into 6 categories. The corresponding road correction coefficients are shown in Table 6. Establish the objective function of fuel cost as formula 19.

\begin{table}[h]
\caption{The solution results of each algorithm on the test set eil101}\label{tab5}%
\begin{tabular}{@{}ccccccc@{}}
\toprule
\textbf{Road Type} & \textbf{Type1}  & \textbf{Type2}  & \textbf{Type3} & \textbf{Type4} 
 & \textbf{Type5} &\textbf{Type6}\\
\midrule
$K_r$	&1.00	&1.10	&1.25	&1.35	&1.45	&1.70\\

\botrule
\end{tabular}
\end{table}

\begin{equation}
    C_1 = p_1 * q * \frac{D}{100} * K_r
\label{eq19}
\end{equation}

(2) Secondly, establish the objective function of personnel cost. This is mainly related to the investigation time. Assuming that the investigator’s driving speed is $v$, the research time spent in each location is $t$, and the investigator’s daily working time is 8 hours, the total number of days $T$ spent by all investigators is expressed by the following formula.

\begin{equation}
    T = \sum_{k=1}^{K} \left[ (\frac{\sum_{i=0}^{N} \sum_{j=0}^{N} x_{ijk} ( \frac{d_{ij}}{v}+t)}{8} ) + 1\right]
\label{eq20}
\end{equation}

Suppose there are m people in each group, and each person's daily salary, meal allowance, and accommodation expenses total $p_2$ yuan. Therefore, the objective function of personnel cost is established as:

\begin{equation}
    C_2 = p_2*T*m
\label{eq21}
\end{equation}

Finally, the objective function of other costs is established, including tolls and parking fees. Assume that the highway toll fee is $p_3$ yuan per kilometer, and the parking fee for parking at each investigation point is set to $p_4$  yuan. The formula is as follows:

\begin{equation}
    C_3=p_3*D+p_4*N
\label{eq22}
\end{equation}

In addition, the mathematical model of this problem also includes some constraints. For example, whether group $k$ goes to site $i$ can be represented by $y_{ki}$; each site can only be investigated by one team; except for the starting point, all investigation locations can only have one out-degree and one in-degree. 

\begin{equation}
    s.t. = \begin{cases}
     \sum_{k=1}^{K}y_{ki} = \begin{cases}
        K & i= 0\\
        1 & i= 1,2,3,...,N
     \end{cases} \\
     \sum_{i=1}^{N}x_{ijk} = y_{kj}  &j=0,1,2,...,N;k = 1,2,...,K\\
     \sum_{j=1}^{N}x_{ijk} = y_{kj}  &j=0,1,2,...,N;k = 1,2,...,K
    \end{cases}
\label{eq23}
\end{equation}

\subsubsection{Model Solving}\label{subsec442}
When solving the problem, the parameters need to be assigned. With reference to the actual situation, the parameter values are as follows: In this example, the 95 gasoline is estimated at 7 yuan per liter; the empty weight comprehensive basic fuel consumption of the selected Volkswagen Sagitar 200TSI vehicle is 7L/100km; according to the highway toll standard of Jiangsu Province, such vehicles are charged 0.45 yuan/km; for the parking fee, we assume that each investigation site will park once and pay 20 yuan; in addition, the personal salary and various subsidies will be set at 250 yuan/day, and one day will be settled if one day is less than one day; Assume that there are 2 members in each group; the time spent at each investigation site is about 1 hour; the average speed of the car on the road is 70 km/h.

\begin{figure}[h]%
\centering
\includegraphics[width=0.65\textwidth]{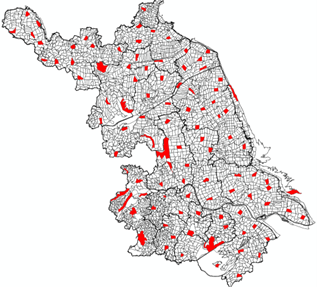}
\caption{Distribution map of selected small watershed}\label{fig9}
\end{figure}

\begin{figure}[h]
  \centering
  \begin{subfigure}[b]{0.24\linewidth}
    \includegraphics[width=\linewidth]{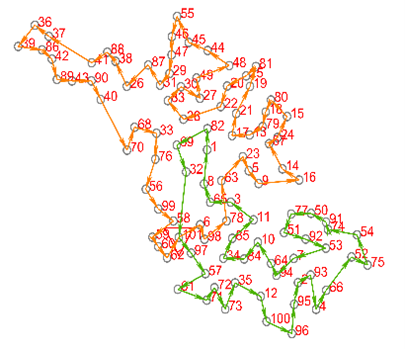}
    \caption{}
    \label{fig:sub1}
  \end{subfigure}
  \begin{subfigure}[b]{0.24\linewidth}
    \includegraphics[width=\linewidth]{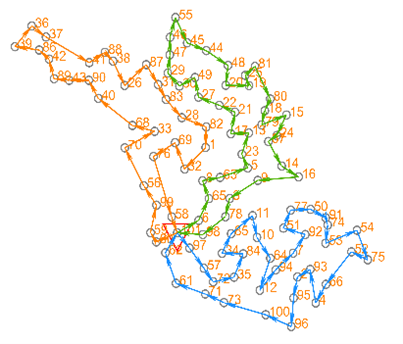}
    \caption{}
    \label{fig:sub2}
  \end{subfigure}
  \begin{subfigure}[b]{0.24\linewidth}
    \includegraphics[width=\linewidth]{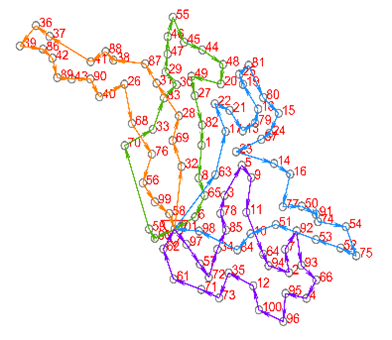}
    \caption{}
    \label{fig:sub3}
  \end{subfigure}
  \begin{subfigure}[b]{0.24\linewidth}
    \includegraphics[width=\linewidth]{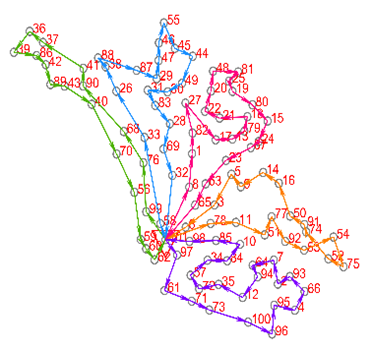}
    \caption{}
    \label{fig:sub4}
  \end{subfigure}
  \caption{Route map of small watershed investigation under different schemes: (a) Divide into 2 groups; (b) Divide into 3 groups; (c) Divide into 4 groups; (d) Divide into 5 groups.}
  \label{fig:fig}
\end{figure}

The center coordinate set of the selected investigation locations corresponds to the city coordinate set used in the Oliver30 and eil101 public datasets. Figure 8 illustrates the randomly selected investigation locations. The experimental results present-ed above provide evidence that the DE-CAFSA algorithm exhibits superior performance across datasets of varying scales. Hence, we devised four schemes for scenarios involving 2, 3, 4, and 5 learning groups. Given the fast convergence of the algorithm, 300 iterations proved sufficient. The trajectory of the research path under different scenarios is depicted in Figure 9. According to the analysis of the data, the least total cost is Scheme 1, and the least time is Scheme 4. The details of these two schemes will be shown in detail below:

\begin{table}[h]
\caption{Research cost analysis table for Scheme 1 (divided into 2 groups)}\label{tab7}%
\begin{tabular}{@{}lcc@{}}
\toprule
\textbf{} & \textbf{Group 1}  & \textbf{Group 2}  \\
\midrule
Number of investigation locations	&60	&40 \\
Group path length (km)	&1866.6778	&1381.9055  \\
Group time spent(hour)	&86.6668	&59.7415 \\
Days of research(day)	&11	&8\\
Total path length(km)    &  \multicolumn{2}{c}{3248.5833}  \\
Fuel cost(CNY)	&914.6721	&677.1337\\
Staff costs(CNY)	&5500	&4000\\
other cost(CNY)	&2040.005	&1421.8575\\
Total cost(CNY)    &  \multicolumn{2}{c}{14553.6683}  \\
\botrule
\end{tabular}
\end{table}

\begin{table}[h]
\caption{Route table of each group in Scheme 1 (divided into 2 groups)}\label{tab8}%
\begin{tabular}{@{}lc@{}}
\toprule
\textbf{Group number} & \textbf{Specific path}  \\
\midrule
Group 1    & \thead{101-6-98-78-63-23-5-9-16-14-67-24-15-80-18-79-13-17-21-19-81-25-20-22-28\\-83-30-27-49-48-44-45-55-46-47-29-31-87-26-38-88-41-37-36-39-86-42-\\89-43-90-40-70-68-33-76-56-99-58-59-60-62-101}      \\
Group 2    & \thead{101-97-57-61-71-72-73-35-12-100-96-95-2-93-4-66-52-75-54-74-91-50-77-\\51-92-53-7-94-64-10-84-34-85-11-3-65-8-1-82-69-32-101}    \\

\botrule
\end{tabular}
\end{table}

\begin{table}[h]
\caption{Research cost analysis table for Scheme 4 (divided into 5 groups)}\label{tab9}%
\begin{tabular}{@{}lccccc@{}}
\toprule
\textbf{} & \textbf{Group 1}  & \textbf{Group 2} & \textbf{Group 3} & \textbf{Group 4} & \textbf{Group 5} \\
\midrule
Number of survey locations	&18	&18	&19	&23	&22\\
Group path length (km)	&706.0736	&786.4963	&775.5737	&801.2243 	&782.0708  \\
Group time spent(hour)	&28.0868	&29.2357	&30.0796	&34.4461	&33.17244 \\
Days of research(day)	&4	&4	&4	&5	&5\\
Total path length(km)    &  \multicolumn{5}{c}{3851.4387}  \\
Fuel cost(CNY)	&345.9761	&385.3832	&380.0311	&392.5999	&383.2147\\
Staff costs(CNY)	&2000	&2000	&2000	&2500	&2500\\
other cost(CNY)	&677.7331	&713.9233	&729.0082	&820.5509	&791.9319\\
Total cost(CNY)    &  \multicolumn{5}{c}{16620.3524}  \\
\botrule
\end{tabular}
\end{table}

\begin{table}[h]
\caption{Route table of each group in Scheme 4 (divided into 5 groups)}\label{tab10}%
\begin{tabular}{@{}lc@{}}
\toprule
\textbf{Group number} & \textbf{Specific path}  \\
\midrule
Group 1    & \thead{101-78-11-51-77-92-53-54-75-52-74-91-50-16-14-9-5-3-6-101}      \\
Group 2    & \thead{101-99-76-68-90-41-37-36-39-86-42-89-43-40-70-56-59-60-62-101}    \\
Group 3    & \thead{101-58-33-26-38-88-87-29-47-46-55-45-44-49-30-31-83-28-69-32-101}    \\
Group 4    & \thead{101-97-61-71-73-100-96-95-4-66-93-2-7-64-94-12-35-72-57-34-84-10-85-98-101}    \\
Group 5    & \thead{101-8-1-82-27-17-13-79-18-21-22-20-48-81-25-19-80-15-24-67-23-63-65-101}    \\
\botrule
\end{tabular}
\end{table}

The experimental results offer a range of optimization schemes, and in our case, Scheme 1 is chosen to minimize the cost. However, if the primary objective is to minimize the time cost of the investigation activity, Scheme 4 can be selected instead.

The DE-CAFSA algorithm has demonstrated notable features such as ease of implementation, fast execution speed, and excellent performance. It has exhibited strong performance not only on public datasets but also when applied to real-world problems. The versatility of DE-CAFSA makes it suitable for a variety of investigation activities, allowing for the maximization of benefits even with limited resources.

\section{Conclusions}\label{sec5}
In this work, we propose a novel approach called the chaotic artificial fish swarm algorithm based on multiple population differential evolution (DE-CAFSA) to address the investigation path planning problem. Our experimental results demonstrate that DE-CAFSA outperforms existing algorithms when applied to the Oliver30 and eil101 public datasets. Moreover, DE-CAFSA provides high-quality solutions that effectively meet the practical requirements of the problem examples presented in this research. From a theoretical perspective, this study offers a viable solution for the multi-traveling salesman problem and introduces a fresh perspective for similar problems, contributing to the advancement of research in this field. From a practical standpoint, employing DE-CAFSA to optimize investigation paths yields shorter time and cost-efficient schemes. This optimization method significantly reduces the cost of investigation activities, enhances the efficiency of the investigation process, and conserves valuable resources. The versatility of DE-CAFSA enables its wide application in various path planning scenarios, providing robust support for different investigation activities.

\backmatter

\bmhead{Data Availability Statement}

The oliver30 data set and eil101 data set in the TSPLIB data set used 678
in this study is openly available from: http://comopt.ifi.uni-heidelberg.de/software/TSPLIB95/.

\bibliography{sn-bibliography.bib}% common bib file

\end{document}